%% file: Manuscript.tex
\documentclass[10pt,journal,compsoc]{IEEEtran}

\ifCLASSOPTIONcompsoc
  \usepackage[nocompress]{cite}
\else
  \usepackage{cite}
\fi
\usepackage{graphicx}
\usepackage[hidelinks]{hyperref}
\usepackage{scalerel}
\usepackage{tikz}
\usetikzlibrary{svg.path}
\usepackage{color}
\usepackage{amsmath}
\usepackage{comment}

\definecolor{orcidlogocol}{HTML}{A6CE39}
\newcommand{\orcidicon}{
    \hspace{-2mm}
  \begin{tikzpicture}
  \draw[orcidlogocol, fill=orcidlogocol] (0,0)
  circle [radius=0.16]
  node[white] {{\fontfamily{qag}\selectfont \tiny ID}};
  \end{tikzpicture}
  \hspace{-2mm}
}

\newcommand{\orcidauthorA}{0000-0003-1015-2115} 
\newcommand{\orcidauthorB}{0009-0002-6755-9880} 
\newcommand{\orcidauthorC}{0000-0000-0000-000Z}

\newcommand{\orcidA}{\href{https://orcid.org/\orcidauthorA}{\orcidicon}}
\newcommand{\orcidB}{\href{https://orcid.org/\orcidauthorB}{\orcidicon}}

\begin{document}

\title{ISS-Geo142: A Benchmark for Geolocating Astronaut Photography from the International Space Station}

\author{Vedika~Srivastava\orcidA{}, Hemant~Kumar~Singh\orcidB{}, and~Jaisal~Singh
  \IEEEcompsocitemizethanks{%
    \IEEEcompsocthanksitem
      The authors are with Boston University, Boston MA 02215, USA.
      E-mail: \texttt{\{vedikas, hks, jaisal\}@bu.edu}.
  }
}

\markboth{}%
{Shell \MakeLowercase{\textit{}}}

\IEEEtitleabstractindextext{%
\begin{abstract}
This paper introduces \textbf{ISS-Geo142}, a curated benchmark for geolocating astronaut photography captured from the International Space Station (ISS). Although the ISS position at capture time is known precisely, the specific Earth locations depicted in these images are typically not directly georeferenced, making automated localization non-trivial. ISS-Geo142 consists of 142 images with associated metadata and manually determined geographic locations, spanning a range of spatial scales and scene types.

On top of this benchmark, we implement and evaluate three geolocation pipelines: a neural network based approach (NN-Geo) using VGG16 features and cross-correlation over map-derived Areas of Interest (AOIs), a Scale-Invariant Feature Transform based pipeline (SIFT-Match) using sliding-window feature matching on stitched high-resolution AOIs, and \textbf{TerraByte}, an AI system built around a GPT-4 model with vision capabilities that jointly reasons over image content and ISS coordinates. On ISS-Geo142, NN-Geo achieves a match for 75.52\% of the images under our evaluation protocol, SIFT-Match attains high precision on structurally rich scenes at substantial computational cost, and TerraByte establishes the strongest overall baseline, correctly geolocating approximately 90\% of the images while also producing human-readable geographic descriptions.

The methods and experiments were originally developed in 2023; this manuscript is a revised and extended version that situates the work relative to subsequent advances in cross-view geo-localization and remote-sensing vision--language models. Taken together, ISS-Geo142 and these three pipelines provide a concrete, historically grounded benchmark for future work on ISS image geolocation.
\end{abstract}

\begin{IEEEkeywords}
Geolocation, VGG16, Convolutional Neural Networks (CNN), Scale-Invariant Feature Transform (SIFT), Template Matching, GPT-4 Vision, Earth Observation, Benchmark.
\end{IEEEkeywords}}

\maketitle

\IEEEdisplaynontitleabstractindextext
\IEEEpeerreviewmaketitle

\ifCLASSOPTIONcompsoc
\IEEEraisesectionheading{\section{Introduction}\label{sec:introduction}}
\else
\section{Introduction}
\label{sec:introduction}
\fi
\IEEEPARstart{A}{stronaut} photography from the International Space Station (ISS) provides unique, high-resolution views of Earth that are valuable for education, outreach, and scientific analysis. However, while the ISS orbit is well characterized and the platform position at capture time is known, the precise ground footprint of each image is typically not directly georeferenced. Automatically recovering the geographic location of these images is therefore a challenging and practically relevant problem.

Most existing geolocation methods and benchmarks focus on ground-level imagery, dense urban scenes, or conventional satellite views. ISS imagery differs in important ways: viewing geometry can be oblique, scales vary widely, and atmospheric effects and cloud cover are common. These factors can cause methods designed for standard cross-view or street-level geo-localization to degrade or fail when applied directly.

In this work, we introduce \textbf{ISS-Geo142}, a curated benchmark of 142 ISS images paired with metadata and manually determined geographic locations. The dataset spans a range of spatial scales, from highly zoomed-in scenes covering less than 1\,km\(^2\) to broad views covering more than 300\,km\(^2\), and includes urban regions, coastlines, river systems, agricultural landscapes, and more. We use ISS-Geo142 to study how three different classes of methods behave on this task and to establish baselines for future work.

Specifically, we implement and compare:
\begin{itemize}
    \item \textbf{NN-Geo}, a neural network based pipeline that uses VGG16 convolutional features and cross-correlation over AOIs constructed from web map tiles centered at the ISS position.
    \item \textbf{SIFT-Match}, a feature-based pipeline built around Scale-Invariant Feature Transform (SIFT) descriptors and sliding-window matching on stitched, high-resolution AOI images.
    \item \textbf{TerraByte}, an AI pipeline that uses GPT-4 with vision capabilities to jointly interpret image content and ISS coordinates, producing both location hypotheses and semantic descriptions.
\end{itemize}

On ISS-Geo142, TerraByte achieves the highest overall accuracy among our pipelines, correctly geolocating approximately 90\% of images under our evaluation protocol, while NN-Geo successfully matches 75.52\% of images and SIFT-Match provides strong, interpretable matches in structurally rich regions at higher computational cost. We treat these pipelines and scores as reference baselines for this benchmark.

\subsection{Contributions}
The main contributions of this paper are:
\begin{itemize}
    \item We present \textbf{ISS-Geo142}, a benchmark dataset of 142 ISS astronaut images with metadata and manually determined geographic locations, designed for systematic evaluation of ISS image geolocation methods.
    \item We develop \textbf{NN-Geo}, a VGG16-based geolocation pipeline that uses convolutional features and cross-correlation over AOIs derived from web map tiles centered at ISS coordinates.
    \item We adapt a \textbf{SIFT-Match} pipeline to large stitched AOIs, analyze its strengths and limitations on ISS imagery, and quantify its computational trade-offs.
    \item We introduce \textbf{TerraByte}, a GPT-4 Vision-based pipeline that uses coordinate-aware prompting to produce both geolocation predictions and semantic geographic descriptions, and we show that it provides the strongest baseline on ISS-Geo142.
    \item We report a comparative evaluation of these three pipelines on ISS-Geo142, highlighting typical success cases and failure modes, and discuss how the work, originally done in 2023, relates to later developments in large-scale astronaut imagery localization and remote-sensing vision--language modeling.
\end{itemize}

Our goal is not to claim a definitive state-of-the-art solution, but to provide a concrete, reproducible benchmark and a set of baselines that future methods can improve upon.

\section{Related Work}
In the evolving landscape of remote sensing and Earth observation, substantial effort has been devoted to geolocating images, particularly for urban and ground-level scenarios. Methods that address ISS or other space-based photography are less common and often tailored to specific mission constraints or metadata availability. Here we highlight representative work that informs our approach and situates it in the broader geo-localization literature.

The work ``Find My Astronaut Photo'' \cite{Stoken_2023_CVPR} leverages astronaut photo metadata to match Earth's surface with corresponding Landsat images, using SE2-LoFTR for feature matching and MAGSAC for robust homography estimation. This pipeline takes advantage of high-quality reference imagery and precise geometry, and demonstrates that astronaut photography can be linked to Earth observation datasets. However, the method focuses on particular image sources and does not fully address the diversity of landscapes and imaging conditions encountered in ISS imagery more broadly.

The study ``Fine-Grained Cross-View Geo-Localization'' \cite{wang2023finegrained} proposes a method to align ground-level images with satellite data using a correlation-aware homography estimator. This approach is designed to handle occlusions, perspective differences, and seasonal variations, and achieves strong performance on benchmark datasets. Applying such techniques directly to ISS imagery is nontrivial, as the viewing geometry, scale, and atmospheric path are quite different from typical cross-view datasets.

Cheng et al.\ introduce ``Image-to-GPS Verification Through A Bottom-Up Pattern Matching Network'' \cite{cheng2018imagetogps}, which performs image-to-GPS verification by matching image features to reference imagery and deciding whether a claimed location is plausible. This approach is highly effective in urban and structured environments, but it assumes the availability of a dense reference database and is most naturally suited to terrestrial settings.

In ``Georeferencing Urban Nighttime Lights Imagery Using Street Network Maps'' \cite{rs14112671}, the authors exploit the structure of street networks to georeference nighttime lights imagery in urban areas. This method illustrates how ancillary vector data can help geolocate images but is focused on urban nighttime imagery and does not directly generalize to the diverse, often rural or oceanic scenes present in ISS photographs.

The work ``Large-scale Image Geo-Localization Using Dominant Sets'' \cite{zemene2017largescale} provides a scalable method for geo-localizing images using reference databases with known GPS coordinates. Dominant Set Clustering is used for robust feature matching and graph-based reasoning. While this approach scales to large image collections, it is primarily designed for terrestrial datasets and benefits from structured features such as roads and buildings, which may appear differently or be less prominent in space-based imagery.

These studies collectively illustrate a spectrum of strategies for image geolocation: metadata exploitation, cross-view alignment, verification frameworks, use of auxiliary vector data, and graph-based clustering. They also highlight important limitations when moving to ISS imagery: varying scale, atmospheric effects, cloud cover, and the lack of dense, directly corresponding reference imagery for arbitrary astronaut photographs. ISS-Geo142 is intended to complement such work by providing a focused benchmark and baselines for this particular regime.

\subsection{Historical Context}
\label{subsec:historical-context}
The methods and experiments presented in this paper were originally developed in 2023, at a time when automated ISS astronaut photography localization and large-scale remote-sensing vision--language models were still emerging. Since then, the field has progressed, with larger astronaut imagery datasets, more powerful cross-view transformers, and remote-sensing-specific vision--language models being introduced.

Within this evolving landscape, ISS-Geo142 and our three pipelines should be viewed as an early, small-scale benchmark and baseline study focused specifically on ISS astronaut photography. The benchmark captures what could be achieved with relatively lightweight CNN and SIFT-based methods, combined with early access to GPT-4 Vision, and provides a historical reference point that later work can surpass and compare against.

\section{Technology Overview of Geolocation Pipelines}
The three geolocation pipelines described in this section combine established computer-vision techniques with geospatial reasoning for ISS imagery. They differ in how they extract and compare features, in their reliance on external APIs or models, and in their computational cost. Together, they define a set of reference baselines on ISS-Geo142.

\subsection{Pipeline 1: Neural Network-Based Geolocation (NN-Geo)}
The NN-Geo pipeline combines convolutional neural network features with geospatial analysis to geolocate images captured by the ISS. At its core, the pipeline uses the VGG16 neural network, a deep convolutional model pre-trained on the ImageNet dataset, which is repurposed here for extracting features from ISS imagery. These features form the basis for matching a query ISS image to a region within a larger AOI image.

Given a query image, NN-Geo first extracts metadata from the EXIF header. Information such as camera model, focal length, and ISS GPS coordinates is obtained when available. The ISS location is used as the center for an AOI, which is fetched using the Mapbox API. We employ a zoom level of 9.5 for AOI images, which provides a balance between spatial detail and coverage, making it feasible to search over the AOI while retaining enough structure for matching.

The AOI bounding box is calculated using the Web Mercator projection, which approximates the Earth's surface and is widely used for web maps. Given the central geographic coordinates of the ISS location \((\phi, \lambda)\) in decimal degrees (latitude and longitude), the map's dimensions in pixels (\(width \times height\)) and the tile size \(T = 512\) pixels (standard for the Mapbox API) are considered. The scale factor at zoom level \(Z\) is \(S = 2^Z \times T\). The latitude is converted to radians and mapped to the Mercator \(Y\) coordinate:
\begin{equation}
Y_{\text{merc}} = \ln\left(\tan\left(\frac{\phi \pi}{180} + \frac{\pi}{4}\right)\right).
\end{equation}
The longitude is similarly converted, and the pixel coordinates \((x_{\text{center}}, y_{\text{center}})\) of the the AOI center are computed as:
\begin{equation}
x_{\text{center}} = \left(\frac{\lambda + 180}{360}\right) \times S,\quad y_{\text{center}} = \left(1 - \frac{Y_{\text{merc}}}{\pi}\right) \times \frac{S}{2}.
\end{equation}
Given these central pixel coordinates and the AOI image width and height, the corner pixel coordinates are determined by simple offsets, and then converted back to geographic coordinates using:
\begin{equation}
    x' = \frac{x}{S} \times 360 - 180
\end{equation}
\begin{equation}
y' = 2 \times \left( \tan^{-1}\left( e^{\left(1 - \frac{2 \cdot y}{S}\right) \cdot \pi} \right) - \frac{\pi}{2} \right) \times \frac{180}{\pi}.
\end{equation}

The core of NN-Geo is the feature extraction and cross-correlation. Both the query image and AOI image (or tiles derived from it) are passed through VGG16 to obtain feature maps. Conceptually, each convolutional layer computes
\begin{equation}
F^{(l)} = \sigma\bigl(K^{(l)} * F^{(l-1)}\bigr),
\end{equation}
where \(F^{(l-1)}\) is the input feature map, \(K^{(l)}\) is the kernel at layer \(l\), and \(\sigma\) is a nonlinearity. We use a suitable intermediate layer as the feature representation for matching.

The similarity between the query feature map \(Q\) and the AOI feature map \(A\) is quantified using cross-correlation:
\begin{equation} 
C(x, y) = \sum_{i,j} \left( Q(i, j) - \bar{Q} \right) \left( A(i+x, j+y) - \bar{A} \right),
\end{equation} 
where \(\bar{Q}\) and \(\bar{A}\) denote mean values. High values of \(C(x, y)\) indicate that the query pattern aligns well with a region of the AOI at offset \((x,y)\). The location of the maximum correlation is taken as the predicted match, which is then mapped back to geographic coordinates using the AOI geometry.

\subsection{Pipeline 2: Scale-Invariant Feature Transform-Based Matching (SIFT-Match)}

The SIFT-Match pipeline is built around the Scale-Invariant Feature Transform (SIFT) algorithm, a widely used technique in computer vision for detecting and matching local features. SIFT is designed to be invariant to changes in scale, rotation, and moderate illumination variations, making it a natural candidate for ISS imagery, which can exhibit different zoom levels and viewing angles.

The pipeline employs a sliding-window strategy, where the SIFT algorithm matches features of the ISS query image against a set of subsampled images derived from a high-resolution AOI. The AOI itself is generated by stitching together multiple Google Maps tiles to form a large composite image.

\textbf{Metadata Extraction:} As in NN-Geo, metadata from the EXIF header (e.g., GPS coordinates, altitude, camera model, focal length) is used to estimate the ISS location and approximate ground footprint at capture time.

\textbf{AOI Image Generation:} A high-resolution AOI image of 1200\,km \(\times\) 1200\,km, centered on the ISS coordinates at capture time, is constructed by stitching Google Maps tiles into an image of size 30{,}000 \(\times\) 30{,}000 pixels. This provides broad coverage for the matching process while retaining enough detail for SIFT features. [We used an open-source repository for this: ``\url{https://github.com/alduinien/GoogleMapDownloader}".]

\textbf{Matching Process:} The AOI is subdivided into overlapping subsampled images whose size is chosen to roughly match the spatial footprint of the ISS image. For each subsample, SIFT keypoints and descriptors are computed, and matched against those of the query image. A matching score (e.g., number of good matches or a related metric) is computed for each subsample, and the subsample with the highest score is selected as the best candidate match. The geographic coordinates of the subsample center provide the predicted location.

\textbf{Enhancements for Accuracy and Efficiency:} To avoid spending computation on images dominated by ocean with little land structure, we include a preclusion criterion to filter out predominantly aquatic regions, in line with the characteristics of our dataset. Additionally, we store the GPS coordinates of the center pixels of the top two best-matched subsamples. This allows us to handle cases where the highest-scoring match is visually similar but geographically incorrect. One such case involves a Manhattan query image where the best match corresponded to a desert region with a similar outline, while the second-best match correctly corresponded to Manhattan, New York City. In such situations, inspecting the second-best candidate can improve effective accuracy.

\textbf{Data Management and Validation:} Results are stored in CSV files to facilitate analysis across different computing environments, including Docker containers. Validation is conducted using images with known coordinates, comparing predicted locations to ground-truth labels and measuring distance errors.

The SIFT-Match pipeline is computationally demanding due to the size of the AOI and the cost of feature extraction and matching. On our setup, processing a single image can take on the order of 15 minutes. Nevertheless, it can achieve high precision in structurally rich regions, especially in urban areas where distinct features such as coastlines, road networks, and river patterns provide strong cues.

\subsection{Pipeline 3: TerraByte - AI-Enhanced Geospatial Analysis}

TerraByte is an AI-enhanced geospatial analysis pipeline that combines ISS imagery with the GPT-4 model equipped with vision capabilities. Unlike NN-Geo and SIFT-Match, which operate primarily on numerical features and pixel-level similarity, TerraByte uses a large vision--language model to interpret the image content and reason about geography in a more semantic way.

TerraByte takes as input an ISS image along with its associated ISS coordinates at capture time. These coordinates provide a coarse prior on the geographic region. The GPT-4 model with vision is then prompted to (i) describe the visible features in the image, and (ii) infer a likely geographic location given the content and the ISS position. Examples of features it may describe include coastlines, peninsulas, urban layouts, river systems, and mountain ranges.

Because GPT-4 has been trained on large-scale textual and visual data, it can leverage prior knowledge about characteristic shapes and configurations of geographic features. For example, when given an image of the Sinai Peninsula and the Nile River in Egypt along with ISS coordinates, TerraByte can produce a description such as:

\textit{``The image shows a distinctive geological feature: a narrow waterway running through an arid region, separating two larger landmasses with varying terrain. The presence of cloud formations near the water indicates that the image was taken from a significant altitude, fitting the perspective of the International Space Station.}

\textit{Given the ISS location at approximately 33.40787°N latitude and 22.99734°E longitude, this places the ISS above Northeast Africa. The waterway shown in the image is very likely the Suez Canal, a man-made canal in Egypt that connects the Mediterranean Sea to the Red Sea, allowing for direct maritime passage between Europe and Asia without navigating around Africa.}

\textit{The image shows the canal running from the top left to the bottom right, with the Sinai Peninsula to the right and the Eastern Desert of Egypt to the left. The northern end of the Red Sea is visible at the bottom right corner, and the Mediterranean Sea would be out of the frame at the top left corner.''}

On ISS-Geo142, TerraByte consistently outperforms NN-Geo and SIFT-Match in terms of the fraction of images correctly geolocated under our evaluation protocol, establishing the strongest baseline among the three pipelines on this benchmark. At the same time, its dependence on a proprietary, subscription-based model and sensitivity to prompt design are important practical limitations.

\vskip 0.2in
Taken together, NN-Geo, SIFT-Match, and TerraByte illustrate three complementary approaches to ISS image geolocation: convolutional feature correlation, local invariant feature matching, and large-scale vision--language reasoning. On ISS-Geo142 they provide a concrete set of baselines for future methods to compare against.

\section{Evaluations}
\subsection{Dataset}
This study uses the \textbf{ISS-Geo142} benchmark, a curated dataset of 142 images captured from the International Space Station (ISS) to support the development and evaluation of geolocation pipelines. The dataset is intentionally diverse in terms of geography, scale, and visual content. As such, our results should be interpreted as specific to this benchmark, but the setup is designed so that future methods can be evaluated in the same way.

\begin{figure}
    \centering
    \includegraphics[width=\linewidth]{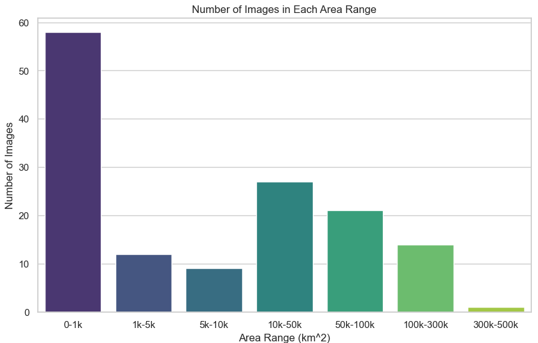}
    \caption{Area range categorization of ISS-Geo142 images. The bar chart shows the distribution of the dataset by surface area coverage on Earth. Area estimates are derived using the field-of-view (FoV) formula, $\text{FoV} = 2 \times \arctan\left(\frac{d}{2f}\right)$, where $d$ is the sensor dimension and $f$ is the focal length of the camera lens. Focal length and camera model are extracted from the image metadata, and sensor dimensions are obtained from camera specifications.}
    \label{fig:area-eval}
\end{figure}

Each image is categorized based on the approximate area it covers on Earth's surface, ranging from highly detailed views covering less than 1\,km\(^2\) to broader scenes spanning more than 300\,km\(^2\). This categorization, visualized in Fig.~\ref{fig:area-eval}, is used to analyze how performance varies with image scale. Images were obtained from astronaut photography sources and publicly available satellite imagery repositories, and include a variety of topographies, land uses, and natural features.

\begin{figure}
    \centering
    \includegraphics[width=\linewidth]{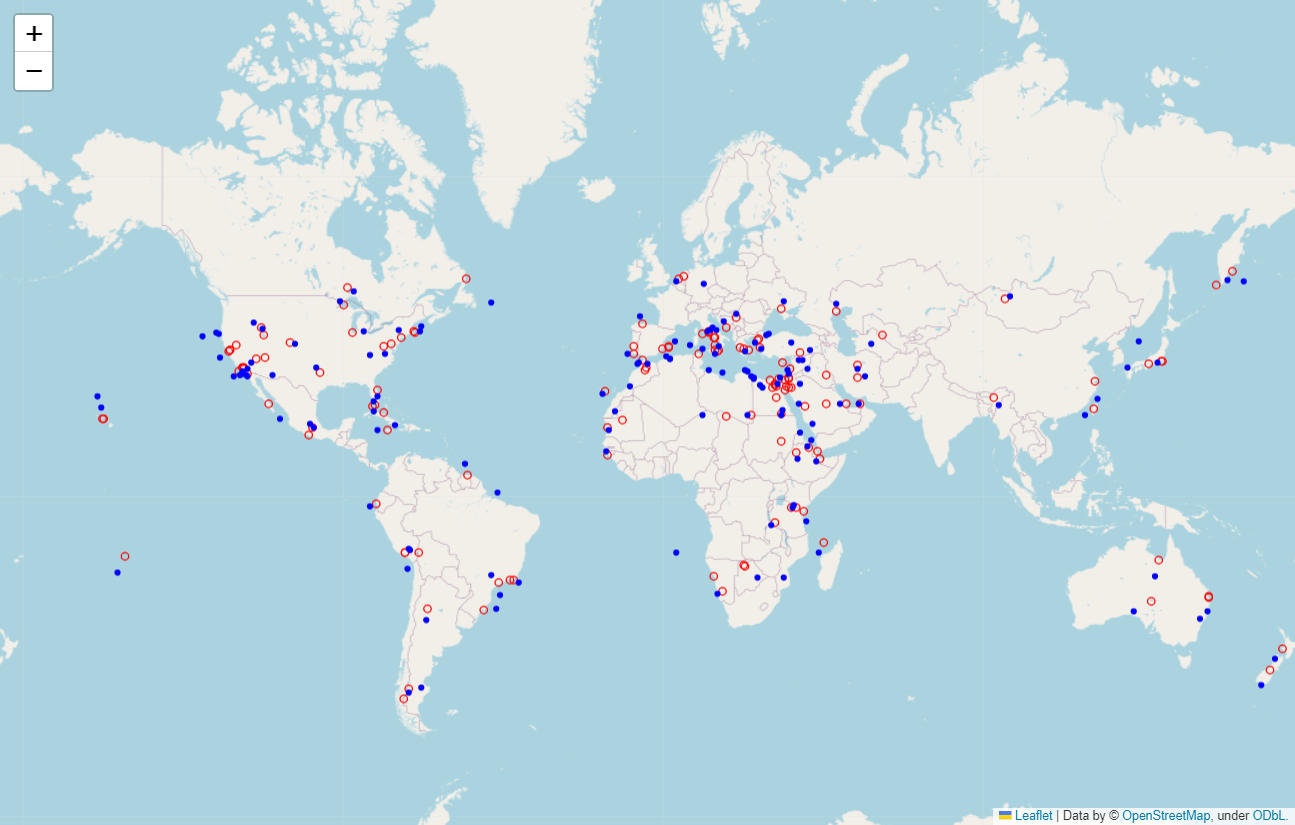}
    \caption{Geospatial distribution of ISS-Geo142. Blue points indicate the ISS location at the time of capture; red points indicate manually geolocated image footprints. The dataset spans a variety of regions and conditions and serves as a testbed for the three geolocation pipelines.}
    \label{fig:dataset-distribution-as-per-location}
\end{figure}

The images are geographically distributed across multiple continents, providing a range of environmental conditions, including urban centers, natural landforms, water bodies, and agricultural regions. Figure~\ref{fig:dataset-distribution-as-per-location} illustrates the global distribution of the dataset. Blue markers indicate the location of the ISS when the image was captured, and red markers denote the manually geolocated image footprint.

Ground-truth labels indicating the actual geographic locations depicted in the images were determined by cross-referencing with online mapping services and, where needed, manual expert inspection. Preprocessing steps included noise reduction, resolution normalization, and basic color correction to standardize the images before passing them to the pipelines.

\subsection{Results}
We evaluate NN-Geo, SIFT-Match, and TerraByte on ISS-Geo142, focusing on approximate success rates, computational characteristics, and suitability for different types of ISS imagery. Table~\ref{tab:combined_pipelines} summarizes the main findings and serves as the reference baseline table for this benchmark.

\begin{table*}[h]
    \centering
    \caption{Comparison and qualitative performance summary of geolocation pipelines on ISS-Geo142. Accuracy values are specific to this benchmark and the evaluation protocol described in the text.}
    \label{tab:combined_pipelines}
    \renewcommand{\arraystretch}{1.5}
    \resizebox{\textwidth}{!}{
    \begin{tabular}{|p{1.5cm}|p{1.2cm}|p{2cm}|p{1.3cm}|p{2.5cm}|p{2.5cm}|p{2.5cm}|p{2.5cm}|}
    \hline
    \textbf{Pipeline} & \textbf{Accuracy} & \textbf{Computational Efficiency} & \textbf{Scalability} & \textbf{Best Use Scenario} & \textbf{Pros} & \textbf{Cons} & \textbf{Future Work} \\ 
    \hline
    NN-Geo & 75.52\% & High & Moderate & Zoomed-in images with clear features and limited cloud cover & Effective on zoomed-in images with clear features; relatively space- and compute-efficient on our setup. & Struggles with heavy cloud cover and varying zoom levels; performance depends on AOI design. & Improve image pre-processing and AOI selection, and explore alternative feature layers or networks. \\ 
    \hline
    SIFT-Match & Variable & Low (resource-intensive, $\sim$15 min per image) & Low & Well-defined terrestrial regions, particularly cities and structured coastlines & High precision in well-structured areas; no training required; interpretable feature matches. & Computationally intensive; less effective on broad landscapes and very zoomed-out views. & Reduce computation via hierarchical search or coarser pre-filtering; improve handling of large-scale scenes. \\ 
    \hline
    TerraByte & $\sim$90\% & Moderate (requires GPT-4 subscription) & High (conceptually; subject to API limits) & Complex images where contextual and semantic understanding is useful & Provides rich geographic descriptions and the highest success rate among our pipelines on ISS-Geo142; can reason about landmarks and patterns. & Depends on proprietary model and subscription; behavior can vary with prompts; struggles on some complex multi-feature images and when ISS coordinates are misleading. & Systematically study prompt design, quantify uncertainty, and investigate open-source vision--language alternatives. \\ 
    \hline
    \end{tabular}
    }
\end{table*}

\subsubsection{Neural Network-Based Geolocation (NN-Geo)}
The NN-Geo pipeline using VGG16 features and cross-correlation achieved encouraging results on ISS-Geo142. Using our evaluation criteria, it produced a correct match for 75.52\% of the images. Qualitatively, NN-Geo performs best on images where the AOI adequately covers the true location and the scene contains distinct patterns such as coastlines, river deltas, or city shapes. It is less effective when the scene is heavily obscured by clouds, when the image is extremely zoomed out, or when the AOI is too large relative to the scene structure.

\begin{figure}
    \centering
    \includegraphics[width=\linewidth]{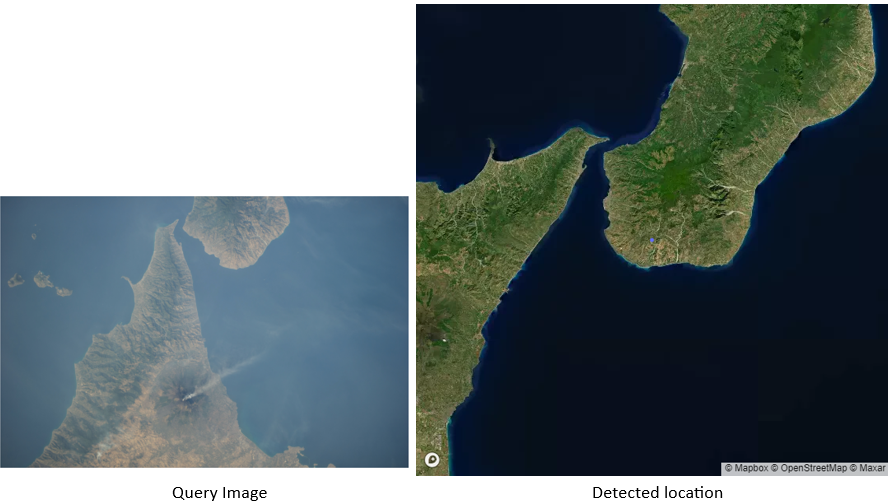}
    \caption{Example NN-Geo result. The left image is the query ISS photograph; the right image shows the predicted location within the AOI, marked with a blue dot. In this case, the NN-based algorithm successfully identifies the corresponding region.}
    \label{fig:NN_result_1}
\end{figure}

Figure~\ref{fig:NN_result_1} shows an example where NN-Geo correctly identifies the location: the left image is the ISS query, and the right image shows the matched region in the AOI, with the predicted location marked.

\subsubsection{Scale-Invariant Feature Transform-Based Matching (SIFT-Match)}
The SIFT-based method focuses on detailed feature matching between the ISS image and subsampled regions from a stitched AOI. On ISS-Geo142, SIFT-Match exhibits strong performance for zoomed-in images with rich structural content (e.g., urban areas or characteristic coastlines), and for cases where the AOI does not include large homogeneous regions.

The trade-off is computational cost: on our hardware, each image requires approximately 15 minutes of processing, making this approach challenging to scale to very large datasets or real-time applications. Performance is also less reliable on very broad or low-texture landscapes.

\begin{figure}
    \centering
    \includegraphics[width=\linewidth]{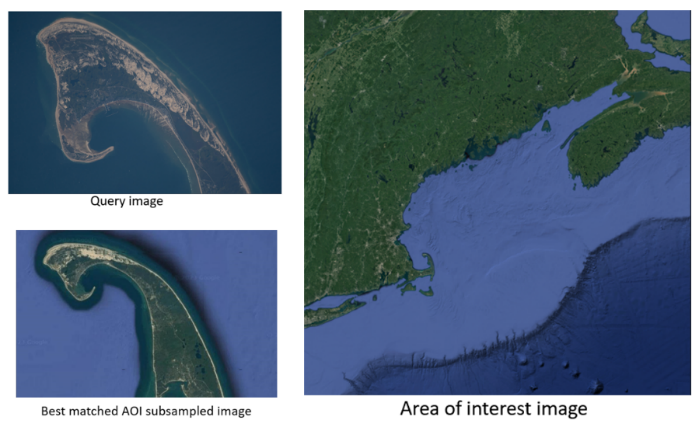}
    \caption{Illustration of the SIFT-based geolocation process. Top left: query ISS image. Right: stitched AOI image from Google Maps tiles. Bottom left: subsampled AOI patch with the highest SIFT matching score, demonstrating successful localization of the query scene.}
    \label{fig:SIFT_result_1}
\end{figure}

Figure~\ref{fig:SIFT_result_1} illustrates the SIFT-based process. The query image (top left) is compared against subsampled AOI images (right), and the best-matched subsample (bottom left) indicates the predicted location.

\subsubsection{TerraByte - AI-Enhanced Geospatial Analysis}
TerraByte, based on GPT-4 Vision, combines approximate ISS coordinates with semantic understanding of the imagery. On ISS-Geo142, TerraByte successfully geolocated approximately 90\% of the images under our evaluation protocol and consistently outperformed NN-Geo and SIFT-Match in terms of success rate. In addition to proposing locations, it provides textual descriptions of the geographic features in the image, which can be useful for interpretation and quality checking.

As noted earlier, TerraByte can struggle when images contain multiple prominent geographic features that could correspond to different regions, or when the ISS coordinates differ significantly from the imaged area (e.g., due to camera orientation or oblique views). In such cases, the model may produce plausible but incorrect location hypotheses.

\begin{figure}
    \centering
    \includegraphics[width=\linewidth]{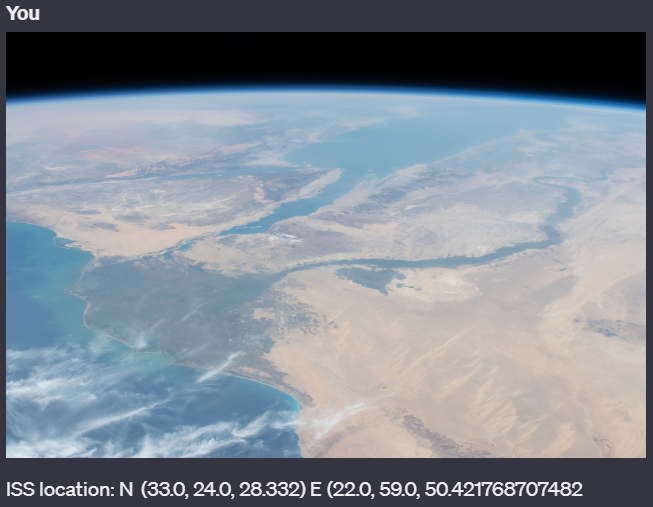}
    \caption{Example prompt used with TerraByte: ISS image (left) and associated ISS coordinates (right) provided as context to GPT-4 Vision. The model responds with both a textual description and a candidate geographic location.}
    \label{fig:terrabyte_prompt_image}
\end{figure}

Despite these limitations, TerraByte currently provides the strongest overall baseline on ISS-Geo142 among the pipelines we implemented, and illustrates the potential of large vision--language models for this task.

\section{Conclusion}
This paper introduced \textbf{ISS-Geo142}, a benchmark for geolocating astronaut photography from the International Space Station, along with three reference pipelines: NN-Geo, SIFT-Match, and TerraByte. Using this 142-image dataset with manually determined ground truth locations, we evaluated how convolutional feature correlation, SIFT-based sliding-window matching, and GPT-4-based vision--language reasoning perform on ISS imagery.

On ISS-Geo142, NN-Geo demonstrates that off-the-shelf convolutional features combined with AOI-based cross-correlation can correctly match 75.52\% of images under our protocol, especially for zoomed-in scenes with clear structure. SIFT-Match shows that classical local feature matching can achieve high precision in structurally rich regions, at the cost of substantial computation. TerraByte, built around GPT-4 Vision, establishes the strongest baseline on this benchmark, correctly geolocating approximately 90\% of images while also providing semantic geographic descriptions.

We do not claim that these pipelines are optimal or that ISS-Geo142 is exhaustive. Instead, we position the dataset and baselines as a starting point for more systematic work on ISS image geolocation. Future research can improve upon these baselines by scaling up the dataset, refining AOI selection and feature representations, developing hierarchical or coarse-to-fine search strategies, and exploring open-source vision--language models as more accessible alternatives to proprietary services.

Because the underlying work was carried out in 2023, ISS-Geo142 and the NN-Geo, SIFT-Match, and TerraByte baselines also serve as a historical snapshot of what was achievable with relatively lightweight methods and early access to GPT-4 Vision. We hope that making this benchmark and its baselines explicit will help future methods situate their progress and foster more cumulative research at the intersection of remote sensing, computer vision, and large-scale AI models for space-based Earth observation.

\ifCLASSOPTIONcompsoc
  \section*{Acknowledgments}
\else
  \section*{Acknowledgment}
\fi
The authors wish to express their sincere gratitude to \textit{\href{https://www.terc.edu/profiles/david-libby/}{Mr. David Libby}}, Chief Technology Officer at \textit{\href{https://www.terc.edu/}{TERC}}, for supporting this project and providing access to relevant resources. We are also grateful for the continuous encouragement that Mr. Libby provided throughout the duration of our work. Our appreciation extends to \textit{\href{https://www.bu.edu/cds-faculty/profile/thomas-gardos/}{Prof. Thomas Gardos}} and \textit{\href{https://www.linkedin.com/in/oindrilla-chatterjee/}{Prof. Oindrilla Chatterjee}}, whose guidance and feedback were instrumental in shaping the research and analysis presented in this paper.

\input{appendix}

\ifCLASSOPTIONcaptionsoff
  \newpage
\fi

\bibliographystyle{IEEEtran}
\bibliography{references}

\end{document}

%% file: appendix.tex
\appendices

\section{Underperforming Methods and Additional Experiments}
\label{app:underperforming}

This appendix documents methods and experiments that we explored during the 2023 development of ISS-Geo142 but ultimately did not adopt as main pipelines. Although these approaches underperformed or were impractical in our setting, we include them here to clarify what was tried and to help future work avoid similar dead ends.

\subsection{LoFTR and SE2-LoFTR}
\label{app:loftr}

We first experimented with LoFTR \cite{sun2021loftr} and SE2-LoFTR \cite{bokman2022se2loftr, e2cnn} for matching ISS photographs to reference imagery. These methods are designed for wide-baseline matching and have shown strong performance on terrestrial benchmarks, but we observed several limitations on ISS-Geo142:

\begin{enumerate}
    \item The models struggled with the very large spatial scale and diversity of Earth topography as seen from orbit.
    \item Many ISS images lack sharp, man-made landmarks; in such cases LoFTR often produced diffuse or unstable correspondences.
    \item Atmospheric haze, clouds, and lighting variations led to noisy matches and frequent failure cases.
\end{enumerate}

\begin{figure}
    \centering
    \includegraphics[width=\linewidth]{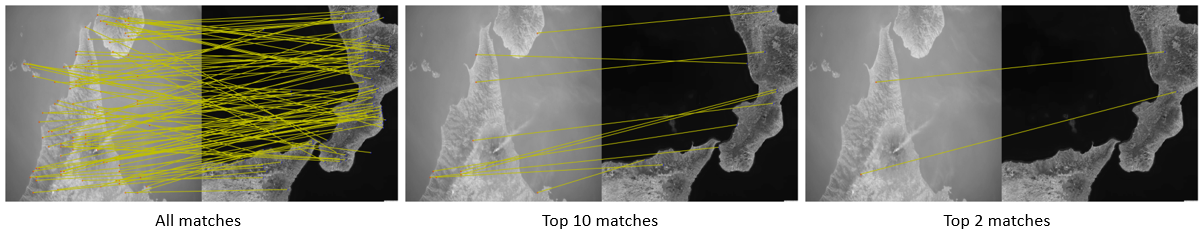}
    \caption{Feature matching with LoFTR between an ISS image and a reference Earth image. From left to right: all matches, top 10 matches, and top 2 matches. Although some correspondences are plausible, many are noisy, and overall performance on ISS-Geo142 was not competitive with the pipelines in the main text.}
    \label{fig:loftr-result-eg}
\end{figure}

Figure~\ref{fig:loftr-result-eg} shows an example with LoFTR: all detected matches (left), the top 10 (middle), and the top 2 (right). While some meaningful correspondences are found, the signal is mixed with substantial noise.

\begin{figure}
    \centering
    \includegraphics[width=\linewidth]{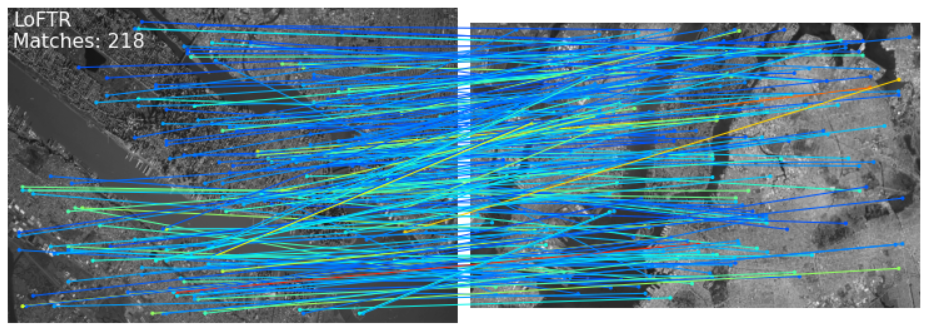}
    \caption{SE2-LoFTR feature matching output between an ISS image and a reference image. Blue and green lines indicate higher- and lower-confidence matches, respectively. On ISS-Geo142, such match sets proved too noisy to yield robust geolocation compared to the baselines reported in the main paper.}
    \label{fig:Se2-loftr_result}
\end{figure}

For SE2-LoFTR, we observed similar behavior. An example is shown in Fig.~\ref{fig:Se2-loftr_result}, where 218 matches are identified but include many low-confidence and spatially inconsistent correspondences that did not translate into reliable geolocation.

\subsection{CasMTR}
\label{app:casmtr}

We also explored CasMTR (Cascade Transformer) for learning spatially informative keypoints. In principle, such architectures could better capture large-scale context in ISS imagery. In practice, two issues prevented us from using CasMTR as a main baseline:

\begin{enumerate}
    \item The model’s GPU and memory requirements were prohibitive in our resource-constrained environment, especially when applied to large AOIs.
    \item Training was difficult to stabilize given the limited amount of labeled data specific to ISS imagery, and performance lagged behind our simpler baselines.
\end{enumerate}

\subsection{Differentiable Spherical Transform}
\label{app:spherical}

We experimented with a differentiable spherical transform to explicitly model Earth curvature when aligning ISS images with reference views. While conceptually appealing, this line of work was not successful on ISS-Geo142 due to:

\begin{enumerate}
    \item The complexity of accurately modeling the viewing geometry for oblique ISS images.
    \item Sensitivity to errors in pose and scale estimation, which often led to misalignment even when the coarse location was correct.
\end{enumerate}

\subsection{Simulated Reference Imagery}
\label{app:simulation}

Another idea was to render synthetic reference images using virtual Earth simulators such as Cesium, and then match ISS images against these simulated views. This approach encountered two main obstacles:

\begin{enumerate}
    \item Rendering large numbers of high-resolution simulated views was too slow for practical use.
    \item The visual gap between simulator renderings and real ISS photographs (texture, lighting, atmosphere) made reliable matching difficult.
\end{enumerate}

\subsection{Automatic Pattern Recognition Heuristics}
\label{app:pattern}

Finally, we implemented lightweight pattern-recognition heuristics (e.g., threshold-based water/land masks, simple coastline detectors) to identify broad geographic structures. These methods were fast but not robust:

\begin{enumerate}
    \item Detection accuracy degraded significantly in the presence of cloud cover or low contrast.
    \item The heuristics often failed to distinguish between visually similar natural features (e.g., different coastlines) or between natural and man-made structures.
\end{enumerate}

\medskip
Overall, these experiments reinforced the difficulty of ISS image geolocation and motivated the focus on the three main pipelines in the paper (NN-Geo, SIFT-Match, and TerraByte). We include this appendix to provide a more complete picture of what was attempted on ISS-Geo142 and to help future work build on, rather than unknowingly repeat, these negative results.